\documentclass[10pt,conference]{IEEEtran}
\IEEEoverridecommandlockouts

\usepackage{cite}
\usepackage{amsmath,amssymb,amsfonts}
\usepackage{graphicx}
\usepackage{textcomp}
\usepackage{xcolor}
\usepackage{pgfplots}
\pgfplotsset{compat=1.18}
\usepackage{array}
\usepackage{algorithm}
\usepackage{algpseudocode}
\usepackage{subcaption}
\usepackage{multirow}
\usepackage{bbm}
\usepackage{url}
\usepackage{tcolorbox}
\usepackage{tabularx}
\usepackage{float}
\usepackage{booktabs}
\usepackage{makecell}
\usepackage{arydshln}
\usepackage{ragged2e}
\usepackage{ulem}
\usepackage{balance}
\usepackage{comment}
\usepackage[utf8]{inputenc}
\usepackage{caption}
\usepackage{tikz}
\usetikzlibrary{trees,positioning,shapes}

\def\BibTeX{{\rm B\kern-.05em{\sc i\kern-.025em b}\kern-.08em
    T\kern-.1667em\lower.7ex\hbox{E}\kern-.125emX}}

\begin{document}

\title{Quantum Entangled Multimodal Fusion Networks (QEMFN): Resource-Aware Hybrid Vision-Language Fusion via Trainable Entanglement}

\author{
Srikar Alla$^{*1}$,
Ali Shiri Sichani$^{1}$,
Chi-Ren Shyu$^{1,2}$\\
$^{1}$Department of Electrical Engineering and Computer Science\\
$^{2}$Institute for Data Science and Informatics\\
University of Missouri, Columbia, MO 65211, USA\\
Email: \{saffw, asp9f, shyuc\}@missouri.edu
}

\maketitle


\begin{abstract}
Multimodal vision--language systems typically fuse image and text embeddings through classical operators such as concatenation, attention, bilinear pooling, or tensor interactions. We propose Quantum Entangled Multimodal Fusion Networks (QEMFN), a hybrid quantum--classical framework that introduces parameterized entanglement as a structured inductive bias for multimodal fusion. Pretrained visual and textual features are projected into compact latent spaces, encoded as angle-parameterized quantum states, processed through intra-modal and paired cross-modal entangling circuits, and measured to produce fused representations for retrieval. Under matched parameter budgets and identical frozen CLIP backbones, QEMFN outperforms classical fusion baselines on COCO-5k and Flickr30k, including multilayer perceptron, tensor fusion, FiLM, cross-attention, compact transformer, and a dequantized paired-topology analogue. An ablation suite isolates the quantum module's contribution from the surrounding classical projections, and quantum-centric analyses report Meyer--Wallach entangling capability, expressibility, gradient variance against barren-plateau bounds, and entropy--performance correlation under controls for training progress alongside an intervention study on the entangling component. QEMFN is executed under shot-based estimation, a noise-modeled fake backend, and a real superconducting device with zero-noise extrapolation. This work does not claim quantum computational advantage; the contribution is the framework together with a controlled empirical and quantum-centric evaluation that positions trainable entanglement as an interpretable, hardware-executable fusion mechanism at scales accessible on contemporary devices.
\end{abstract}




\section{Introduction}

Multimodal vision-language systems jointly model image and text data and underpin applications including image retrieval, semantic search, visual question answering, recommendation, and human-computer interaction. Large-scale contrastive models such as CLIP and ALIGN demonstrated that aligned visual and textual representations can be learned effectively from paired data through contrastive objectives over large-scale supervision, and the resulting embeddings now serve as reusable backbones for many downstream tasks~\cite{radford2021clip, jia2021align}. Transformer-based multimodal architectures including ALBEF, BLIP, and ViLT~\cite{li2021albef, li2022blip, kim2021vilt}, together with token-level systems such as LXMERT, UNITER, and VisualBERT~\cite{tan2019lxmert, chen2019uniter, li2019visualbert}, have further advanced cross-modal understanding through learned alignment modules and attention-based interactions, establishing the dominant paradigm for vision-language modeling. Despite this progress, fusion in most multimodal systems still relies on classical operators including feature concatenation, multilayer perceptrons, bilinear pooling, tensor fusion, and cross-attention~\cite{multimodal_survey}. Capturing higher-order interactions typically requires deeper networks, larger attention stacks, or explicit interaction tensors, increasing memory use and optimization complexity. Many practical fusion layers also impose linear, low-rank, or factorized assumptions that may not align with the structure of cross-modal dependencies. As multimodal systems are increasingly deployed in settings requiring fine-grained alignment and compositional understanding, the design space of fusion mechanisms with different structural priors than standard classical operators becomes worth exploring. This is not a claim that classical fusion is fundamentally limited, but rather a motivation to study alternative inductive biases under controlled conditions and matched parameter budgets.

Hybrid quantum-classical learning has emerged as one such alternative, supported by software platforms including Qiskit and PennyLane~\cite{qiskit, pennylane}. Quantum states evolve in tensor-product Hilbert spaces and support entanglement, in which subsystems exhibit dependencies that cannot be decomposed into independent unimodal components~\cite{biamonte2017quantum, schuld2021ml, raimond2001manipulating, wootters1998entanglement}. Recent hybrid systems have explored quantum components for perception and robotics tasks~\cite{tamilarasan2025qasvs}, and a smaller body of work has investigated quantum approaches to multimodal fusion specifically. From a multimodal perspective, the natural inductive bias is to model visual and textual representations as coupled subsystems within a shared quantum state, with cross-modal interactions implemented through parameterized entangling operations rather than through attention or tensor products. The empirical question this study addresses is whether such a structured entangling module, evaluated under controlled conditions including parameter-matched classical baselines, ablation of the quantum component, and execution on real quantum hardware, provides a measurable and interpretable contribution to multimodal fusion at scales accessible on contemporary devices. To address this question, we introduce Quantum Entangled Multimodal Fusion Networks (QEMFN), a hybrid framework in which pretrained visual and textual embeddings are projected into compact latent spaces, encoded as angle-parameterized quantum states, processed through structured intra-modal and paired cross-modal entangling layers, and measured to produce fused representations for retrieval. Pretrained encoders remain frozen during training so that comparisons isolate the contribution of the fusion stage. The framework is evaluated on COCO-5k and Flickr30k under image-text retrieval, text-image retrieval, and cross-dataset transfer settings~\cite{lin2014coco, young2014image}, together with quantum-centric analyses of circuit resources, expressibility, gradient behavior, entanglement dynamics, noise robustness, and execution on contemporary superconducting hardware with shot-based estimation and zero-noise extrapolation.

This work does not claim quantum computational advantage. The base configuration uses 12 logical qubits and shallow circuit depth, which is classically simulable at the scales studied here. The contribution is the framework itself together with a controlled empirical and quantum-centric evaluation. Specifically, this paper contributes the following. First, a hybrid fusion module that models cross-modal interactions through paired entanglement and is evaluated under matched parameter budgets against classical alternatives spanning multilayer perceptron, low-rank bilinear, tensor fusion, FiLM, cross-attention, compact transformer, and a dequantized paired-topology baseline. Second, an ablation suite that isolates the contribution of the quantum module from the surrounding classical projections, addressing the concern that performance gains in hybrid quantum-classical systems may be driven entirely by classical components. Third, quantum-centric analyses including empirical Meyer-Wallach entangling capability, expressibility, gradient variance compared against barren-plateau reference bounds~\cite{mcclean2018barren}, and entropy-performance correlation studies that control for training progress, alongside an intervention study that perturbs the entangling component directly. Fourth, hardware execution under shot-based estimation, noise-modeled backend simulation, and a real superconducting device with zero-noise extrapolation, characterizing the gap between ideal simulation and current device performance~\cite{ibmq_noise}.


\section{Related Work}

Multimodal learning has traditionally combined modality-specific features through classical fusion operators including concatenation, bilinear pooling, tensor fusion, gating, and low-rank factorization~\cite{multimodal_survey}. These approaches established the foundations of cross-modal reasoning and retrieval but face tradeoffs between expressivity and efficiency: explicit interaction tensors can be costly, while heavily factorized variants reduce capacity. Compact alternatives such as FiLM-style conditioning and gated residual fusion are widely used in lightweight retrieval pipelines but draw their interaction structure from the same classical operator vocabulary. Recent progress has been driven by large-scale pretrained vision-language models. CLIP and ALIGN learn aligned image and text representations through contrastive objectives over large paired datasets~\cite{radford2021clip, jia2021align}, producing transferable embeddings used as backbones across downstream tasks. Dual-stream and encoder-decoder systems including ALBEF, BLIP, and ViLT improve multimodal understanding through learned alignment modules and attention-based interactions~\cite{li2021albef, li2022blip, kim2021vilt}, while LXMERT, UNITER, and VisualBERT model token-level dependencies between image regions and language tokens through cross-attention~\cite{tan2019lxmert, chen2019uniter, li2019visualbert, vaswani2017attention}. The fusion stages in these systems remain fundamentally classical.

Quantum machine learning has explored variational quantum circuits, quantum kernels, and hybrid quantum-classical optimization as components of learning systems~\cite{biamonte2017quantum, schuld2021ml, havlivcek2019supervised, quantum_kernel, boyapally2025qsannviq}. Parameterized quantum circuits typically act as trainable feature maps while classical networks handle preprocessing and prediction. Theoretical studies suggest that such circuits can model structured correlations in tensor-product Hilbert spaces with compact parameterizations~\cite{schuld2021ml}, though practical realization depends on circuit design, observable choice, and trainability characteristics including the barren-plateau phenomenon~\cite{mcclean2018barren}. The relationship between quantum representations and classical surrogates remains active, with results such as Tang's dequantization~\cite{quantum_advantage_ml} showing that some quantum constructions admit efficient classical analogues at near-term scales. A smaller body of work has investigated quantum approaches to multimodal fusion specifically~\cite{qu2023qmfnd, chen2024qmcl}, together with hybrid quantum systems for related perception tasks~\cite{tamilarasan2025qasvs}. These studies have shown that quantum components can contribute to multimodal classification, but evaluations have generally been limited to small-scale tasks, treated quantum components as auxiliary rather than central modules, or omitted controls that isolate the quantum contribution from surrounding classical scaffolding.

QEMFN addresses these gaps by positioning a parameterized entangling circuit as the central fusion mechanism rather than as an auxiliary component, and by evaluating it under controls that isolate its contribution. The framework is compared against parameter-matched classical baselines including a dequantized paired-topology analogue, ablated to separate the quantum module from its classical scaffolding, characterized through quantum-centric diagnostics including Meyer-Wallach entangling capability and gradient variance against barren-plateau bounds, and executed on real superconducting hardware to characterize the gap between simulation and device performance.


\begin{figure*}[t]
\centering
\includegraphics[width=\textwidth]{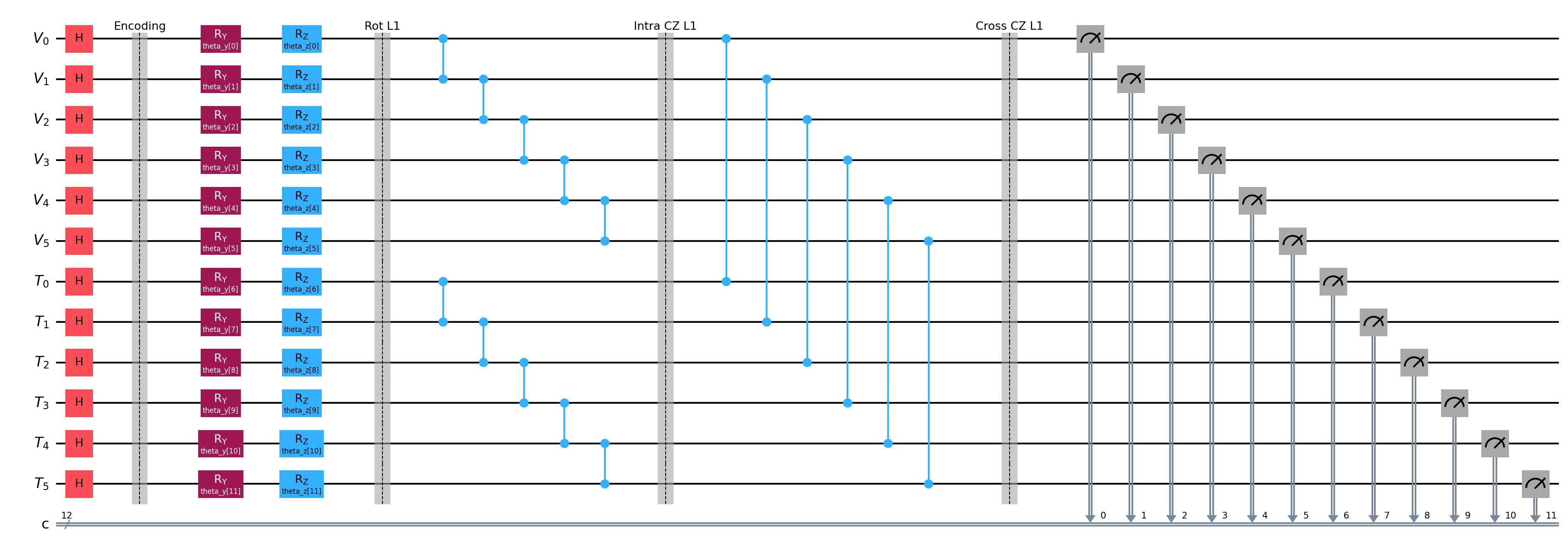}
\caption{QEMFN quantum fusion circuit with paired cross-modal entanglement. Visual qubits $V_0$--$V_5$ and textual qubits $T_0$--$T_5$ are prepared from projected modality embeddings, followed by trainable single-qubit rotations, intra-modal entangling layers, and paired cross-modal couplings $V_i \leftrightarrow T_i$. Final measurements generate the fused representation used for retrieval.}
\label{fig:qemfn_circuit}
\end{figure*}


\section{QEMFN Framework}

\subsection{System Overview}

In the vision-language setting, the input is an image $x$ and a paired text description $y$, and the objective is to map both into a shared space where matched pairs are close and mismatched pairs are separated. Most existing systems achieve this through classical fusion operators such as concatenation, attention, or bilinear interactions. QEMFN instead formulates fusion as a parameterized quantum process in which visual and textual features are embedded into a shared quantum state and structured dependencies are learned through trainable entanglement. The architecture has four stages: feature extraction using frozen pretrained encoders, projection into compact latent spaces, quantum encoding with trainable entangling evolution, and measurement into a fused classical representation. Figure~\ref{fig:qemfn_circuit} illustrates the pipeline. Given a pair $(x, y)$, the system outputs a fused representation $\mathbf{z}_{\mathrm{out}}$ and similarity score $s(x, y)$.

\subsection{Quantum State Preparation}

Let $\mathbf{v}\in\mathbb{R}^{d_v}$ and $\mathbf{t}\in\mathbb{R}^{d_t}$ denote pretrained visual and textual embeddings produced by frozen encoders. These are projected into quantum-compatible latent vectors and normalized to unit $\ell_2$-norm,
\begin{equation}
\hat{\mathbf{v}} = \frac{W_V\mathbf{v}}{\|W_V\mathbf{v}\|_2}, \qquad \hat{\mathbf{t}} = \frac{W_T\mathbf{t}}{\|W_T\mathbf{t}\|_2},
\end{equation}
where $W_V \in \mathbb{R}^{n_q \times d_v}$ and $W_T \in \mathbb{R}^{n_q \times d_t}$ are trainable. Each normalized component controls a single-qubit rotation under angle encoding, and the initial joint state is the tensor product
\begin{equation}
|\Psi_0\rangle = \bigotimes_{k=1}^{n_q}R_y(\pi\hat{v}_k)|0\rangle \otimes \bigotimes_{k=1}^{n_q}R_y(\pi\hat{t}_k)|0\rangle.
\end{equation}
In the default configuration, 768-dimensional visual and 512-dimensional textual embeddings from CLIP ViT-B/32 are projected into $n_q = 6$ qubits per modality, yielding a 12-qubit register. Angle encoding is used in the main experiments because of its efficient state preparation; amplitude encoding and data re-uploading are evaluated as ablations.

\subsection{Entangling Evolution and Measurement}

The trainable evolution is parameterized as a stack of $L$ layers. Each layer combines trainable single-qubit rotations on every qubit, intra-modal CZ couplings within each modality register, and paired cross-modal CZ gates between matched qubits $V_i \leftrightarrow T_i$:
\begin{equation}
U_\ell(\theta_\ell) = U_{\mathrm{cross}}^{(\ell)} \, U_{\mathrm{intra}}^{(\ell)} \, U_{\mathrm{rot}}^{(\ell)}(\theta_\ell),
\end{equation}
where $U_{\mathrm{rot}}^{(\ell)}(\theta_\ell) = \bigotimes_{k=1}^{2n_q} R_y(\theta_{\ell,k})$. The full evolution is $|\Psi_{VT}\rangle = U_L \cdots U_1 |\Psi_0\rangle$, and the joint density operator and reduced visual state are $\rho = |\Psi_{VT}\rangle\langle\Psi_{VT}|$ and $\rho_V = \mathrm{Tr}_T(\rho)$. The paired cross-modal topology uses $n_q$ inter-modal gates per layer rather than the $\mathcal{O}(n_q^2)$ gates required for fully connected coupling, supporting controlled circuit depth and trainability.

\begin{table*}[t]
\centering
\caption{Image-to-text retrieval on COCO-5k over ten random seeds, mean $\pm$ standard deviation. All methods use the same frozen CLIP ViT-B/32 backbone. ALBEF and BLIP are end-to-end pretrained reference systems and are not parameter-matched. The $p$-value column reports paired Wilcoxon signed-rank tests against QEMFN with Bonferroni correction across the parameter-matched comparisons.}
\label{tab:coco_i2t_main}
\begin{tabular}{lcccccc}
\toprule
Model & Params & R@1 & R@5 & MRR & Latency (ms) & $p$-value \\
\midrule
CLIP (frozen) & -- & $58.2 \pm 0.3$ & $81.4 \pm 0.2$ & $0.691 \pm 0.004$ & 1.4 & -- \\
ALBEF (reference) & 210M & $72.4 \pm 0.2$ & $91.1 \pm 0.1$ & $0.812 \pm 0.003$ & -- & -- \\
BLIP (reference) & 224M & $78.1 \pm 0.2$ & $94.0 \pm 0.1$ & $0.861 \pm 0.002$ & -- & -- \\
\midrule
CLIP + MLP Fusion & 0.42M & $61.7 \pm 0.6$ & $84.0 \pm 0.4$ & $0.721 \pm 0.007$ & 3.8 & $<0.001$ \\
Low-Rank Bilinear Fusion & 0.45M & $63.9 \pm 0.5$ & $85.1 \pm 0.4$ & $0.736 \pm 0.006$ & 5.4 & $<0.001$ \\
Tensor Fusion & 0.48M & $64.8 \pm 0.5$ & $85.9 \pm 0.3$ & $0.744 \pm 0.006$ & 8.9 & $<0.001$ \\
CLIP + Trainable Head & 0.43M & $66.8 \pm 0.5$ & $87.9 \pm 0.3$ & $0.772 \pm 0.006$ & 6.1 & $<0.01$ \\
FiLM / Gated Fusion & 0.41M & $67.3 \pm 0.4$ & $88.4 \pm 0.3$ & $0.771 \pm 0.005$ & 5.8 & $<0.01$ \\
Cross-Attention Fusion & 0.42M & $67.8 \pm 0.4$ & $88.7 \pm 0.3$ & $0.776 \pm 0.005$ & 7.2 & $<0.01$ \\
Compact Transformer Fusion & 0.44M & $68.1 \pm 0.5$ & $88.9 \pm 0.3$ & $0.778 \pm 0.006$ & 9.4 & $<0.05$ \\
Dequantized Paired-Topology Fusion & 0.41M & $68.2 \pm 0.4$ & $89.0 \pm 0.3$ & $0.779 \pm 0.005$ & 6.7 & $<0.05$ \\
QEMFN w/o Entanglement & 0.39M & $67.1 \pm 0.4$ & $88.2 \pm 0.3$ & $0.768 \pm 0.005$ & 11.8 & $<0.01$ \\
QEMFN & 0.41M & $68.9 \pm 0.4$ & $89.4 \pm 0.3$ & $0.781 \pm 0.004$ & 12.5 & -- \\
\bottomrule
\end{tabular}
\end{table*}

Cross-modal coupling is quantified through the von Neumann entropy and its qubit-normalized form,
\begin{equation}
S(\rho_V) = -\mathrm{Tr}(\rho_V \log_2 \rho_V), \qquad S_{\mathrm{norm}} = \frac{S(\rho_V)}{n_V}.
\end{equation}
The fused representation is a vector of measured observable expectations $\mathbf{z} = [\langle O_1 \rangle, \dots, \langle O_M \rangle]$. In the default configuration, $M = 24$, comprising 12 local Pauli-$Z$ observables, six paired cross-modal correlators $Z_{V_i} \otimes Z_{T_i}$, and six intra-modal correlators $Z_{V_i} \otimes Z_{V_{i+1}}$. Thus, $M=24$ follows directly from the selected readout design: one local observable for each of the 12 qubits, one paired correlator for each of the six visual--textual qubit pairs, and six selected intra-modal correlators. This configuration provides complete local coverage together with balanced cross-modal and intra-modal interaction measurements while limiting the shot and readout cost. The value of $M$ is therefore a measurement-design hyperparameter rather than a theoretically optimal constant. Local Pauli-$Z$ observables admit unbiased shot-based estimation on hardware and produce stronger gradients than global cost functions; The observable set was selected to balance interaction coverage and measurement efficiency rather than from a theoretical optimality argument. The fused vector is passed through a linear output head $W_{\mathrm{out}}$ to produce $\mathbf{z}_{\mathrm{out}} = W_{\mathrm{out}}\mathbf{z}$, and similarity between two pairs is the cosine similarity $s_{ij} = \mathbf{z}_{\mathrm{out}}^{(i)} \cdot \mathbf{z}_{\mathrm{out}}^{(j)} / (\|\mathbf{z}_{\mathrm{out}}^{(i)}\| \, \|\mathbf{z}_{\mathrm{out}}^{(j)}\|)$.

\subsection{Optimization and Computational Complexity}

For a minibatch of $N$ pairs, training uses a symmetric InfoNCE objective,
\begin{equation}
\mathcal{L} = -\frac{1}{2N}\sum_{i=1}^{N}\log\frac{\exp(s_{ii}/\tau)}{\sum_{j}\exp(s_{ij}/\tau)} - \frac{1}{2N}\sum_{i=1}^{N}\log\frac{\exp(s_{ii}/\tau)}{\sum_{j}\exp(s_{ji}/\tau)},
\end{equation}
with temperature $\tau$. Only the projections $W_V, W_T$, entangling parameters $\{\theta_\ell\}_{\ell=1}^{L}$, and output head $W_{\mathrm{out}}$ are optimized; pretrained encoders remain frozen. Optimization uses Adam with separate learning rates for classical and quantum parameters to account for the distinct gradient magnitudes typical of variational circuits, and sensitivity to $\tau$ is reported in Section~IV.

The computational footprint is governed by three quantities: the qubit count per modality $n_q$, the circuit depth $L$, and the number of observables $M$. Increasing $n_q$ enlarges the joint Hilbert space to dimension $2^{2n_q}$ and improves representational capacity but increases runtime and on-hardware decoherence exposure. Increasing $L$ enables richer interactions but compounds gate-error accumulation and reduces gradient magnitudes consistent with the barren-plateau analysis in Section~V. Increasing $M$ expands the fused representation but raises shot-budget requirements during hardware execution. The base configuration of $n_q = 6$, $L = 4$, and $M = 24$ is chosen as the empirical balance point characterized in Section~V. At training time the model encodes pairs jointly under full pairwise scoring; retrieval-time deployment supports both this protocol and dual-encoder operation with optional top-$K$ quantum reranking, with corresponding complexity tradeoffs quantified in Section~IV.


\section{Experimental Results}

\subsection{Retrieval Performance}

The primary evaluation is conducted on the COCO-5k benchmark using the standard Karpathy split with a 5k candidate retrieval pool. All compared methods use the same frozen CLIP ViT-B/32 backbone, so differences reflect the fusion mechanism rather than feature quality. The corresponding circuit and observable definitions are given in Section~III. Table~\ref{tab:coco_i2t_main} reports image-to-text retrieval performance under matched parameter budgets, together with paired Wilcoxon signed-rank tests against QEMFN with Bonferroni correction across the parameter-matched comparisons.

QEMFN achieves the strongest Recall@1, Recall@5, and Mean Reciprocal Rank among the parameter-matched fusion baselines. The classical baselines span a progression of interaction structures, including nonlinear feature mixing (MLP), pairwise multiplicative interactions (low-rank bilinear), higher-order feature composition (tensor fusion), feature-wise modulation (FiLM), and attention-based fusion (cross-attention and compact transformer). The closest classical analogue is the dequantized paired-topology fusion, which mirrors the structural connectivity of QEMFN through Kronecker-style mixing in classical space. QEMFN improves over this analogue by 0.7 Recall@1, supporting the interpretation that the entangling layer contributes representational structure beyond what the topology alone provides. This difference is not interpreted as evidence of quantum computational advantage, but as an empirical distinction under matched structural and parameter constraints. The performance gap between QEMFN and its non-entangled variant further isolates the contribution of cross-modal CZ couplings.

Although larger pretrained systems such as ALBEF and BLIP achieve higher absolute performance, they rely on substantially larger end-to-end architectures and extensive paired pretraining. QEMFN focuses on improving the fusion stage while keeping backbone encoders fixed, supporting modular deployment and lightweight adaptation. The latency column shows that the quantum module introduces moderate simulator overhead relative to classical baselines but remains within an order of magnitude of trained classical heads at the base configuration. Performance scales monotonically with effective batch size up to 256, consistent with the role of negative sample diversity in InfoNCE training, and saturates beyond that point.

Table~\ref{tab:coco_t2i} reports text-to-image retrieval on the same benchmark. The relative ordering of methods is consistent with the image-to-text setting, indicating that the fusion structure transfers symmetrically across query directions.

\begin{table}[t]
\centering
\caption{Text-to-image retrieval on COCO-5k over ten seeds.}
\label{tab:coco_t2i}
\begin{tabular}{lccc}
\toprule
Model & R@1 & R@5 & MRR \\
\midrule
CLIP (frozen) & $37.8 \pm 0.3$ & $62.4 \pm 0.3$ & $0.501 \pm 0.005$ \\
CLIP + Trainable Head & $46.7 \pm 0.5$ & $70.6 \pm 0.4$ & $0.586 \pm 0.006$ \\
QEMFN w/o Entanglement & $45.1 \pm 0.4$ & $69.4 \pm 0.3$ & $0.571 \pm 0.005$ \\
QEMFN & $48.4 \pm 0.4$ & $72.0 \pm 0.3$ & $0.603 \pm 0.005$ \\
\bottomrule
\end{tabular}
\end{table}

\subsection{Cross-Dataset Generalization}

To evaluate robustness beyond the training distribution, models trained on COCO-5k are evaluated on Flickr30k without task-specific retraining. Following standard protocol, Flickr30k uses a 1k candidate pool while COCO-5k uses a 5k pool, so absolute scores are not directly comparable across datasets. Table~\ref{tab:flickr_transfer} reports cross-dataset transfer performance.

QEMFN achieves the highest Recall@1 among the compared fusion methods and maintains its lead across Recall@5 and Recall@10. The improved transfer performance suggests that the learned interactions capture relationships extending beyond dataset-specific co-occurrence statistics. While frozen CLIP exhibits a slightly smaller relative drop from in-domain to out-of-domain, this reflects the absence of any fitted alignment rather than a robustness advantage; QEMFN produces higher overall retrieval accuracy in both domains after training. The gap between QEMFN and the non-entangled variant remains stable under distribution shift, indicating that the entangling component does not overfit to dataset-specific patterns.

\begin{table}[t]
\centering
\caption{Cross-dataset transfer from COCO-5k to Flickr30k without task-specific retraining, ten seeds. Flickr30k uses a 1k candidate pool per standard protocol.}
\label{tab:flickr_transfer}
\begin{tabular}{lccc}
\toprule
Model & R@1 & R@5 & R@10 \\
\midrule
CLIP (frozen) & $52.1 \pm 0.3$ & $77.3 \pm 0.2$ & $86.0 \pm 0.1$ \\
CLIP + Trainable Head & $58.7 \pm 0.5$ & $81.2 \pm 0.3$ & $88.8 \pm 0.2$ \\
QEMFN w/o Entanglement & $58.0 \pm 0.4$ & $81.0 \pm 0.3$ & $88.6 \pm 0.2$ \\
QEMFN & $60.8 \pm 0.4$ & $83.2 \pm 0.3$ & $90.1 \pm 0.2$ \\
\bottomrule
\end{tabular}
\end{table}

\subsection{Ablation Study and Contribution Analysis}

A potential concern in hybrid quantum-classical models is whether observed gains arise from the quantum module itself or from the surrounding classical projections. In QEMFN, the visual and textual projections together account for the majority of trainable parameters, while the quantum circuit contributes only 96 trainable rotation angles. To isolate the contribution of each component, Table~\ref{tab:quantum_contribution} reports an ablation suite that varies which components are trainable and which are present.

The first row reports a 96-parameter classical multilayer perceptron operating on the same projected features, providing a parameter-matched classical baseline for the quantum module alone. The quantum-only head, in which projections are frozen at random initialization but the quantum circuit is trained, exceeds this baseline by 2.3 Recall@1, indicating that the quantum circuit contributes representational structure beyond a comparably-sized classical head. Allowing the projection layers to be trainable while replacing the quantum circuit with a fixed random circuit recovers most of the gain from learned projections, but still falls below configurations with trainable quantum parameters. QEMFN improves over the random fixed circuit by 3.1 Recall@1, attributable to the trainable quantum module, and over the non-entangled variant by 1.8 Recall@1, attributable specifically to the cross-modal entangling gates. This decomposition suggests that gains arise from a combination of trainable projection alignment and structured quantum interaction, with each component contributing measurably under controlled comparison.

\begin{table}[t]
\centering
\caption{Ablation isolating the contribution of the quantum module on COCO-5k Recall@1. The first row provides a 96-parameter classical baseline matched to the quantum circuit's parameter count.}
\label{tab:quantum_contribution}
\begin{tabular}{lccc}
\toprule
Variant & Proj. Trainable & Q Params & R@1 \\
\midrule
96-param Classical MLP & No & 0 & $60.4 \pm 0.6$ \\
Quantum-only Head & No & 96 & $62.7 \pm 0.5$ \\
Frozen Projection + QEMFN & No & 96 & $64.1 \pm 0.5$ \\
Random Fixed Circuit & Yes & 0 & $65.8 \pm 0.5$ \\
QEMFN w/o Entanglement & Yes & 96 & $67.1 \pm 0.4$ \\
QEMFN & Yes & 96 & $68.9 \pm 0.4$ \\
\bottomrule
\end{tabular}
\end{table}

\begin{table}[t]
\centering
\caption{Sensitivity of QEMFN retrieval performance to the InfoNCE temperature $\tau$ on COCO-5k.}
\label{tab:temperature_ablation}
\begin{tabular}{ccc}
\toprule
Temperature $\tau$ & R@1 & R@5 \\
\midrule
0.03 & $67.2 \pm 0.6$ & $88.1 \pm 0.4$ \\
0.05 & $68.4 \pm 0.5$ & $89.0 \pm 0.3$ \\
0.07 & $68.9 \pm 0.4$ & $89.4 \pm 0.3$ \\
0.10 & $68.7 \pm 0.4$ & $89.2 \pm 0.3$ \\
0.20 & $67.5 \pm 0.5$ & $88.4 \pm 0.4$ \\
\bottomrule
\end{tabular}
\end{table}

\subsection{Hyperparameter Sensitivity and Retrieval Protocol}

Two practical concerns affect the interpretation of retrieval results. The first is sensitivity to the InfoNCE temperature parameter $\tau$, which controls the sharpness of the contrastive distribution and is known to be sensitive in vision-language training. Table~\ref{tab:temperature_ablation} sweeps $\tau$ across a moderate range. Performance is stable around $\tau = 0.07$, with degradation below 0.05 and above 0.10 reflecting the standard tradeoff between underfit and overfit in contrastive temperature selection. The chosen value is therefore not a knife-edge configuration.

The second concern is retrieval-time computational complexity. Table~\ref{tab:retrieval_complexity} clarifies the protocols available for QEMFN at deployment. Under full pairwise scoring, each candidate requires a fresh joint quantum evaluation, leading to per-query cost that scales with the candidate pool size. Under dual-encoder retrieval, no quantum calls are required at inference and standard approximate nearest neighbor search applies. The intermediate top-$K$ reranking protocol uses dual-encoder retrieval to retain $K$ candidates, then applies the quantum module only to those candidates, decoupling fusion expressivity from candidate-pool size. The experiments in this section use full pairwise scoring at evaluation time to characterize the underlying mechanism, with $K$-reranking deployment left as a practical option. Performance on substantially larger datasets is not established by the present experiments. Computationally, full pairwise evaluation scales linearly with the candidate pool as $\mathcal{O}(NMS)$; therefore, deployment at larger scale would use classical dual-encoder retrieval followed by quantum reranking of a small top-$K$ candidate set. The present hardware results should consequently be interpreted as feasibility characterization rather than evidence of retrieval-time speedup.

\begin{table}[t]
\centering
\caption{Retrieval protocols and per-query computational complexity. $N$ denotes candidate pool size, $K$ the rerank window, $M$ the number of measured observables, and $S$ the number of shots per observable.}
\label{tab:retrieval_complexity}
\setlength{\tabcolsep}{4pt}
\small
\begin{tabular}{lcc}
\toprule
Protocol & Q-Calls & Complexity \\
\midrule
Full Pairwise Scoring & $N$ & $\mathcal{O}(NMS)$ \\
Dual Encoder Only & 0 & $\mathcal{O}(N)$ ANN \\
Top-$K$ Reranking & $K$ & $\mathcal{O}(N) + \mathcal{O}(KMS)$ \\
\bottomrule
\end{tabular}
\end{table}


\section{Quantum-Centric Analysis}

\subsection{Circuit Resources and Expressibility}

A practical advantage of the proposed framework is that its computational structure can be analyzed explicitly through quantum resource metrics, rather than only through downstream task accuracy. Unlike classical multimodal systems whose complexity is distributed across deep stacks of layers, QEMFN concentrates fusion behavior within a compact circuit defined by qubit count, circuit depth, and the number of measured observables. This enables direct study of how architectural scaling affects representational capacity, optimization, and downstream performance. The observable set is selected to balance measurement efficiency and cross-modal interaction capture rather than theoretical optimality.

Expressibility characterizes the ability of the circuit to represent diverse quantum states. Let $U(\theta)$ denote the parameterized circuit and $|\Psi(\theta)\rangle = U(\theta)|0\rangle$. Expressibility is approximated using the Hilbert-Schmidt distance between the induced state distribution and the Haar distribution,
\begin{equation}
E = \mathbb{E}_{\theta}\left[\left\|\rho(\theta) - \rho_{\mathrm{Haar}}\right\|_2^2\right],
\end{equation}
where $\rho(\theta) = |\Psi(\theta)\rangle\langle\Psi(\theta)|$ and lower values indicate closer agreement with the Haar-induced distribution. The entangling capability is quantified using the Meyer-Wallach metric,
\begin{equation}
Q = 2\left(1 - \frac{1}{n}\sum_{k=1}^{n}\mathrm{Tr}(\rho_k^2)\right),
\end{equation}
where $\rho_k$ is the reduced density matrix of the $k$-th qubit and larger values indicate stronger multi-qubit entanglement.

Table~\ref{tab:resource_metrics_qe} summarizes representative configurations. The base configuration achieves moderate Meyer-Wallach entangling capability of approximately 0.61 and expressibility well below the Haar limit at 0.09, which is consistent with the design goal of a structured rather than fully random unitary. The intra-modal CZ couplings together with paired cross-modal CZ gates produce non-trivial entanglement while avoiding the gate growth of fully connected bipartite circuits. Table~\ref{tab:qubit_ablation} examines the effect of qubit allocation on retrieval. Performance increases substantially up to 6+6 qubits and saturates beyond, with marginal gains at 8+8 and 10+10 not justifying the increased simulator latency. The base configuration of 12 logical qubits and 4 layers therefore represents a favorable operating point and lies within the scale of contemporary superconducting and trapped-ion devices, although practical deployment requires transpilation, qubit mapping, shot-based estimation, and error mitigation.

\begin{table}[t]
\centering
\caption{Quantum resource scaling with empirical Meyer-Wallach entangling capability and expressibility, computed over 1000 random parameter samples per configuration.}
\label{tab:resource_metrics_qe}
\begin{tabular}{lcccccc}
\toprule
Configuration & Qubits & Layers & Params & CZ Gates & $Q_{\mathrm{MW}}$ & $E$ \\
\midrule
Small  & 8  & 3 & 48  & 21 & 0.42 & 0.18 \\
Base   & 12 & 4 & 96  & 44 & 0.61 & 0.09 \\
Large  & 16 & 4 & 128 & 60 & 0.73 & 0.05 \\
XLarge & 20 & 5 & 200 & 95 & 0.81 & 0.03 \\
\bottomrule
\end{tabular}
\end{table}

\begin{table}[t]
\centering
\caption{Effect of qubit allocation on retrieval performance and simulator latency. The 6+6 configuration is used as the base.}
\label{tab:qubit_ablation}
\begin{tabular}{lccc}
\toprule
Qubit Configuration & R@1 & MRR & Latency (ms) \\
\midrule
2+2 & 59.8 & 0.701 & 7.0 \\
4+4 & 64.2 & 0.739 & 9.8 \\
6+6 & 68.9 & 0.781 & 12.5 \\
8+8 & 69.3 & 0.784 & 25.8 \\
10+10 & 69.4 & 0.786 & 41.3 \\
\bottomrule
\end{tabular}
\end{table}

\subsection{Trainability and Barren Plateau Analysis}

Optimization in variational quantum circuits is constrained by the barren plateau phenomenon, in which gradient variance decreases exponentially with system size. For circuits with global cost functions and depth growing with qubit count, the variance scales as $\mathcal{O}(2^{-n})$, rendering gradient-based optimization infeasible. Cerezo et al.~\cite{cerezo2021cost} showed that this scaling can be relaxed for shallow circuits with local observables, where the relevant bound is determined by the support of the cost function rather than the full Hilbert-space dimension. Table~\ref{tab:gradient_variance_bp} reports empirical gradient variance for QEMFN at increasing scale, alongside the global and local barren-plateau reference bounds. At the base configuration, empirical variance of $1.2 \times 10^{-3}$ exceeds the global bound by approximately five orders of magnitude and tracks the local-cost bound, which is the regime predicted for shallow circuits with local observables and local cost functions~\cite{cerezo2021cost}. Empirical variance decreases with scale and remains comparable to the local bound for the Large and XLarge configurations, which is consistent with the diminishing returns observed in Table~\ref{tab:qubit_ablation} and motivates the choice of the base configuration as the operating point. The use of local Pauli-Z observables rather than global cost functions, together with shallow circuit depth and structured paired connectivity, places QEMFN within a trainable regime distinct from the worst-case barren plateau scaling.

\begin{table}[t]
\centering
\caption{Empirical gradient variance compared against global and local barren-plateau (BP) reference bounds. Empirical values are computed via parameter-shift gradients over 1000 random initializations.}
\label{tab:gradient_variance_bp}
\setlength{\tabcolsep}{3pt}
\small
\begin{tabular}{lccccc}
\toprule
Config. & Qubits & Depth & Empirical & Global BP & Local BP \\
\midrule
Small  & 8  & 3 & 5.8e-3 & 1.5e-5  & 4.6e-3 \\
Base   & 12 & 4 & 1.2e-3 & 3.6e-8  & 1.3e-3 \\
Large  & 16 & 4 & 2.7e-4 & 8.9e-11 & 3.7e-4 \\
XLarge & 20 & 5 & 5.4e-5 & 2.2e-13 & 1.0e-4 \\
\bottomrule
\end{tabular}
\end{table}

\subsection{Entanglement Dynamics and Noise Robustness}

Entanglement dynamics provide insight into how multimodal interactions evolve during training. The cross-modal coupling is quantified using the von Neumann entropy of the reduced visual subsystem,
\begin{equation}
S(\rho_V) = -\mathrm{Tr}(\rho_V \log_2 \rho_V), \qquad S_{\mathrm{norm}} = \frac{S(\rho_V)}{n_V},
\end{equation}
where $\rho_V$ is obtained by tracing out the text register from the joint state. The von Neumann entropy reported here is computed from the full simulated statevector. On physical hardware, reduced-state observables, subsystem purities, and entropy estimates may be obtained using classical-shadow or randomized-measurement protocols, although the required measurement budget increases with subsystem size~\cite{huang2020predicting}.

\begin{figure}[t]
\centering
\includegraphics[width=\linewidth]{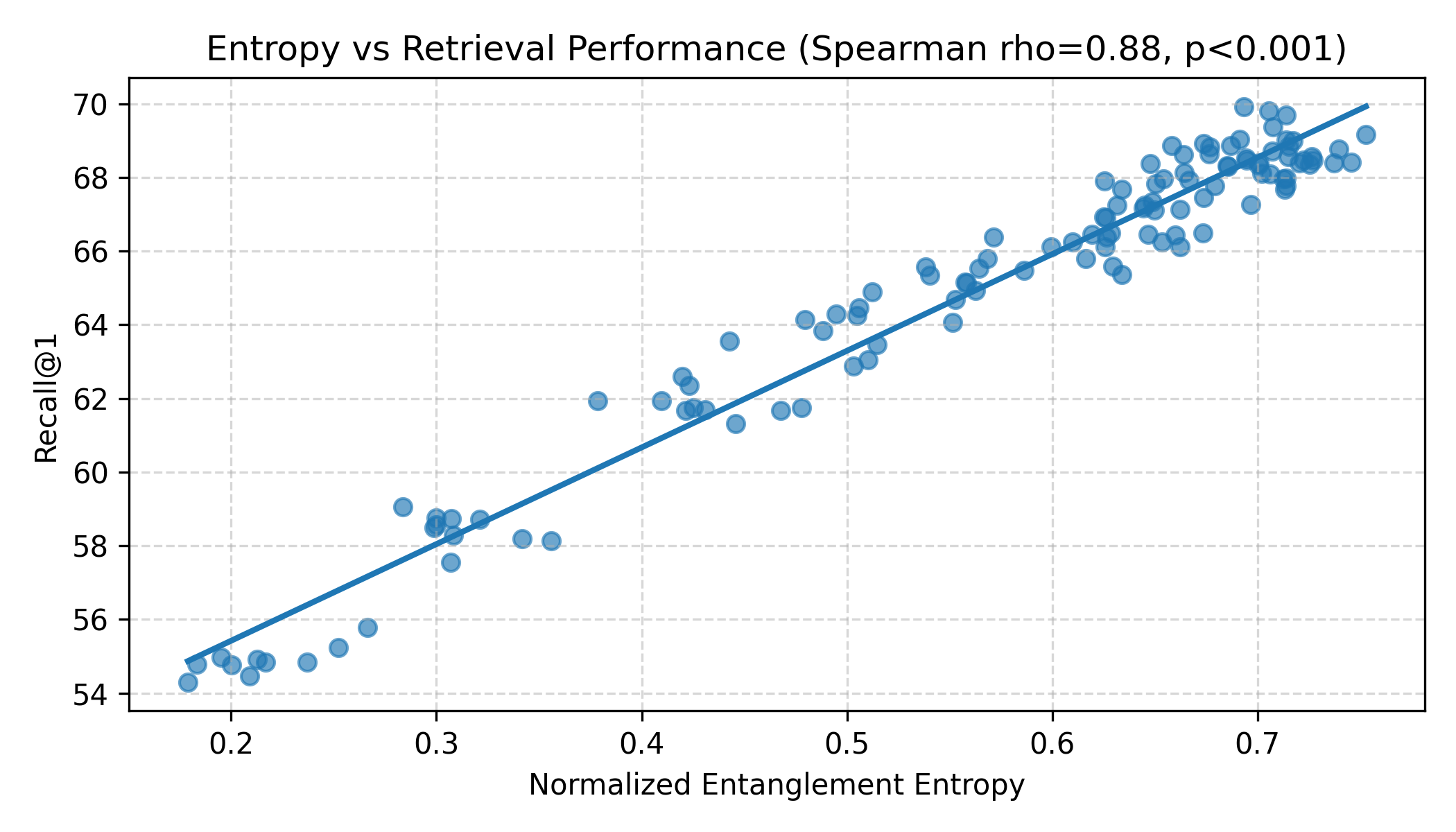}
\caption{Correlation between normalized entanglement entropy and Recall@1 across 12 epochs and 10 random seeds. Each point corresponds to one seed-epoch observation.}
\label{fig:entropy_corr}
\end{figure}

Figure~\ref{fig:entropy_corr} shows the relationship between normalized entanglement entropy and Recall@1 across all 12 training epochs and 10 random seeds, yielding 120 observations. The Spearman rank correlation is $\rho = 0.88$ with $p < 0.001$ and 95\% bootstrap confidence interval $[0.82, 0.93]$. Because both entropy and retrieval performance increase monotonically during training, this raw correlation could in principle reflect a common dependence on training progress. To address this, Table~\ref{tab:entropy_causality} reports correlation analyses controlling for training epoch and validation loss, alongside two classical coupling diagnostics for comparison. The epoch-detrended partial Spearman correlation drops to 0.56, and the validation-loss-controlled correlation to 0.49, both remaining statistically significant. Classical coupling measures including centered kernel alignment and the cross-covariance spectrum produce weaker correlations, suggesting that entanglement entropy carries information beyond what these classical surrogates capture.

\begin{table}[t]
\centering
\caption{Entropy-performance analysis after controlling for training progress, alongside classical coupling diagnostics for comparison.}
\label{tab:entropy_causality}
\begin{tabular}{lcc}
\toprule
Analysis & Correlation & $p$-value \\
\midrule
Raw Spearman $\rho(S_{\mathrm{norm}}, \mathrm{R@1})$ & 0.88 & $<0.001$ \\
Epoch-detrended partial Spearman & 0.56 & $<0.01$ \\
Controlled for validation loss & 0.49 & $<0.05$ \\
Classical CKA vs. R@1 & 0.41 & $<0.05$ \\
Cross-covariance spectrum vs. R@1 & 0.38 & $<0.05$ \\
\bottomrule
\end{tabular}
\end{table}

To move beyond correlation, Table~\ref{tab:entanglement_intervention} reports an intervention study that directly perturbs the entangling component. Freezing the entangling gates at initialization reduces both entanglement entropy and Recall@1, with retrieval falling to 67.4. Replacing the structured paired connectivity with random cross-modal pairings reduces both quantities further, with Recall@1 falling to 66.9, indicating that the alignment between physical qubit pairs and learned representations matters beyond mere connectivity. Removing cross-modal gates entirely eliminates the entanglement growth signature while leaving Recall@1 close to the non-entangled baseline at 67.1, confirming that the gate-removal ablation in Table~\ref{tab:quantum_contribution} reflects the absence of cross-modal coupling. Regularizing the model toward low entanglement during training reduces Recall@1 to 66.5, providing direct evidence that suppressing the quantity correlated with performance also suppresses performance.

\begin{table}[t]
\centering
\caption{Intervention study testing whether cross-modal entanglement affects retrieval performance.}
\label{tab:entanglement_intervention}
\begin{tabular}{lcc}
\toprule
Intervention & $S_{\mathrm{norm}}$ & R@1 \\
\midrule
Full QEMFN & $0.71 \pm 0.02$ & $68.9 \pm 0.4$ \\
Frozen Entangling Gates & $0.58 \pm 0.03$ & $67.4 \pm 0.5$ \\
Random Cross-Modal Pairing & $0.53 \pm 0.03$ & $66.9 \pm 0.5$ \\
Cross-Modal Gates Removed & $0.31 \pm 0.02$ & $67.1 \pm 0.4$ \\
Entropy-Regularized Low Coupling & $0.42 \pm 0.03$ & $66.5 \pm 0.5$ \\
\bottomrule
\end{tabular}
\end{table}

Robustness to hardware noise is evaluated under a depolarizing channel applied to single-qubit and two-qubit operations,
\begin{equation}
\mathcal{E}_p^{(1)}(\rho) = (1-p)\rho + \frac{p}{2}I, \qquad \mathcal{E}_p^{(2)}(\rho) = (1-p)\rho + \frac{p}{4}I_4,
\end{equation}
where $p$ is the depolarizing probability. Table~\ref{tab:noise_sensitivity} reports retrieval and entanglement entropy at $p \in \{0, 10^{-3}, 5\times10^{-3}, 10^{-2}\}$. Performance remains stable under mild noise and degrades gradually as $p$ increases. Entanglement entropy decreases monotonically with noise, indicating that depolarization weakens cross-modal coupling, and the joint degradation of entropy and retrieval reinforces the link between the two quantities established in the intervention study.

\begin{table}[t]
\centering
\caption{Noise sensitivity of the base QEMFN configuration under a depolarizing channel.}
\label{tab:noise_sensitivity}
\begin{tabular}{lccc}
\toprule
Noise Level $p$ & R@1 & MRR & $S_{\mathrm{norm}}$ \\
\midrule
0 & 68.9 & 0.781 & 0.71 \\
$10^{-3}$ & 68.1 & 0.773 & 0.69 \\
$5\times10^{-3}$ & 65.7 & 0.748 & 0.64 \\
$10^{-2}$ & 62.4 & 0.711 & 0.58 \\
\bottomrule
\end{tabular}
\end{table}

\subsection{Preliminary Hardware Evaluation}

To evaluate the framework beyond ideal simulation, the base configuration was executed under shot-based simulation, on a noise-modeled fake IBM backend, and on a real superconducting device. Table~\ref{tab:hardware_aware} reports the resulting Recall@1 and Mean Reciprocal Rank across execution modes, together with the transpiled depth and CX gate count after routing to the device coupling map and basis gate set $\{\mathrm{cx}, u\}$ at optimization level 3. Performance degrades from 68.9 under ideal statevector simulation to 68.2 under 4096-shot estimation, reflecting the variance of finite-shot expectation values. Execution on the noise-modeled fake backend further reduces Recall@1 to 65.9, and execution on the real superconducting device at 8192 shots yields 63.8. Application of zero-noise extrapolation recovers approximately one Recall@1 point, reaching 65.1. The remaining gap between ideal simulation and mitigated hardware execution reflects the cumulative effect of finite-shot estimation, two-qubit gate errors at the deployed transpiled depth, and decoherence over circuit execution time. The transpiled depth on hardware is roughly five times the logical depth due to coupling-map routing and basis-gate decomposition, and the CX gate count approximately doubles relative to the logical CZ count. These results establish that QEMFN executes on contemporary superconducting hardware and that error mitigation provides measurable recovery, while also quantifying the gap between simulation-based evaluation and current device performance. This work does not claim quantum computational advantage; the framework is evaluated as a structured hybrid fusion mechanism whose hardware behavior is characterized rather than relied upon for the central results. At the current scale, the circuit remains classically simulable; the focus of this work is on structured hybrid modeling rather than asymptotic quantum advantage.

\begin{table*}[t]
\centering
\caption{Preliminary hardware evaluation of the base QEMFN configuration under shot-based estimation, noise-modeled backend simulation, and real superconducting hardware execution. Transpiled depth and CX count reflect routing to the device coupling map and basis gate set at optimization level 3.}
\label{tab:hardware_aware}
\begin{tabular}{lccccc}
\toprule
Execution Mode & Shots & Transpiled Depth & CX Count & R@1 & MRR \\
\midrule
Ideal Statevector & -- & 41 & 44 (CZ logical) & $68.9 \pm 0.4$ & $0.781 \pm 0.004$ \\
Shot Simulation & 1024 & 118 & 132 & $67.6 \pm 0.6$ & $0.769 \pm 0.006$ \\
Shot Simulation & 4096 & 118 & 132 & $68.2 \pm 0.5$ & $0.775 \pm 0.005$ \\
Fake IBM Backend & 4096 & 284 & 211 & $65.9 \pm 0.8$ & $0.748 \pm 0.008$ \\
Real IBM Backend & 8192 & 196 & 148 & $63.8 \pm 1.1$ & $0.721 \pm 0.012$ \\
ZNE-Mitigated Hardware & 8192 & 196 & 148 & $65.1 \pm 0.9$ & $0.736 \pm 0.010$ \\
\bottomrule
\end{tabular}
\end{table*}


\section{Discussion}

The empirical results position QEMFN as a structured hybrid fusion module rather than a quantum-advantaged learning system. Across image–text retrieval, cross-dataset transfer, and parameter-matched comparisons, QEMFN consistently outperforms compact classical fusion baselines under identical frozen backbones. The closest classical analogue, a dequantized paired-topology fusion that mirrors the circuit connectivity in classical space, achieves 68.2 Recall@1 compared to 68.9 for QEMFN. This gap is modest and should be interpreted as an empirical difference under matched structural constraints rather than evidence of quantum computational advantage. The contribution of this work is a controlled demonstration that paired entanglement, combined with classical projections, provides measurable improvements while exposing interpretable signals through quantum-specific diagnostics. Ablation studies show that performance gains arise from the interaction between projection alignment and structured quantum evolution. A 96-parameter classical multilayer perceptron applied to the same projected features achieves 60.4 Recall@1, while a trained quantum circuit with frozen projections achieves 62.7, indicating additional representational structure beyond a parameter-matched classical head. The full model improves over a random fixed circuit by 3.1 Recall@1 due to trainable quantum parameters, and over the non-entangled variant by 1.8 Recall@1 due to cross-modal entanglement. These results address the concern that improvements in hybrid quantum-classical systems may be driven entirely by classical components.

Quantum-centric analysis provides further structure beyond retrieval accuracy. The Meyer–Wallach entangling capability of approximately 0.61 and expressibility of 0.09 indicate a structured, non-random unitary, consistent with the paired cross-modal topology. Empirical gradient variance aligns with the local-cost barren plateau regime described by Cerezo et al., rather than the global regime, confirming that the model operates within a trainable region. Although diminishing returns are observed at larger qubit counts, this behavior is consistent with known optimization challenges in variational quantum circuits. The relationship between entanglement and performance is supported by both correlation and intervention analyses. Partial Spearman correlations of 0.56 and 0.49, controlling for training epoch and validation loss, exceed classical coupling diagnostics such as centered kernel alignment. Intervention experiments in which entangling gates are frozen, cross-modal pairings are randomized, or low entanglement is enforced consistently reduce both entropy and Recall@1, indicating that suppressing entanglement also suppresses performance. Hardware experiments characterize the gap between ideal simulation and current quantum devices. Recall@1 decreases from 68.9 under ideal statevector simulation to 68.2 under shot-based estimation, 65.9 on a noise-modeled backend, and 63.8 on real superconducting hardware at 8192 shots. Zero-noise extrapolation recovers approximately one Recall@1 point. This degradation reflects finite-shot variance, two-qubit gate errors, and decoherence at transpiled depths near 200 two-qubit operations. The observed decrease in entanglement entropy under increasing depolarizing noise further supports the link between coherence and the entanglement-driven retrieval signal. These findings are consistent with prior studies showing that noise and two-qubit gate errors constrain the effective depth of near-term quantum circuits and motivate error-mitigation methods~\cite{ibmq_noise}.

This work does not claim quantum computational advantage. Circuits of the size considered here remain classically simulable and admit classical analogues such as the dequantized paired-topology baseline reported in Table~\ref{tab:coco_i2t_main}. The present comparisons do not rule out stronger tensor-network-inspired, random-feature, or further-optimized compact classical surrogates, and the reported gain should therefore be interpreted only within the evaluated baseline set. A formal analysis of dequantizability through tensor-network methods or complexity-theoretic arguments is left for future work \cite{quantum_advantage_ml}. The contribution of this study lies in the structured hybrid fusion mechanism, the isolation of the quantum module's contribution from its classical scaffolding, and the characterization of trainability, entanglement dynamics, and hardware behavior. These elements support QEMFN as an interpretable and resource-characterized fusion module under controlled conditions.


\section{Conclusion and Future Work}

This work introduced Quantum Entangled Multimodal Fusion Networks (QEMFN), a hybrid quantum-classical framework that uses parameterized entanglement as an inductive bias for multimodal fusion. Pretrained visual and textual features are projected into compact latent spaces, encoded as angle-parameterized quantum states, processed through intra-modal and paired cross-modal entangling layers, and measured to produce fused representations for retrieval. The contribution is the framework together with a controlled empirical and quantum-centric evaluation, not a claim of quantum computational advantage. Experiments on COCO-5k and Flickr30k show that QEMFN outperforms parameter-matched classical fusion baselines including multilayer perceptron, tensor fusion, FiLM, cross-attention, compact transformer, and a dequantized paired-topology analogue under identical frozen CLIP backbones. An ablation suite isolates the contribution of the quantum module from the surrounding classical projections, and quantum-centric analyses report Meyer-Wallach entangling capability, expressibility, gradient variance against barren-plateau bounds, partial entropy-performance correlation controlling for training progress, and an intervention study on the entangling component. Hardware execution under shot-based estimation, noise-modeled backend simulation, and a real superconducting device with zero-noise extrapolation characterizes the gap between ideal simulation and current device performance. Future directions include broader hardware evaluation with shadow tomography for on-hardware entanglement diagnostics, formal dequantizability analysis through tensor-network bond dimension, extension to multimodal reasoning benchmarks such as visual question answering, adaptive circuit architectures, alternative encoding strategies, and distributed multimodal systems built on quantum communication infrastructures~\cite{hasan2023qcs}.


\section*{Acknowledgment}
The computation for this work was performed on the University of Missouri's Quantum Innovation Center, in partnership with IBM Quantum and facilitated by Research Support Services at the University of Missouri, Columbia, MO.



\begin{thebibliography}{00}


\bibitem{radford2021clip}
A.~Radford \emph{et al.}, ``Learning transferable visual models from natural language supervision,'' 
in \emph{Proc. Int. Conf. Mach. Learn. (ICML)}, 2021, pp. 8748--8763.

\bibitem{jia2021align}
C.~Jia \emph{et al.}, ``Scaling up visual and vision-language representation learning with noisy text supervision,'' 
in \emph{Proc. Int. Conf. Mach. Learn. (ICML)}, 2021, pp. 4904--4916.


\bibitem{li2019visualbert}
L. H. Li, M. Yatskar, D. Yin, C.-J. Hsieh, and K.-W. Chang, 
``VisualBERT: A Simple and Performant Baseline for Vision and Language,'' 
\textit{arXiv preprint arXiv:1908.03557}, 2019. 
doi: 10.48550/arXiv.1908.03557.


\bibitem{multimodal_survey}
T.~Baltru{\v{s}}aitis, C.~Ahuja, and L.-P.~Morency, ``Multimodal machine learning: A survey and taxonomy,'' 
\emph{IEEE Trans. Pattern Anal. Mach. Intell.}, vol. 41, no. 2, pp. 423--443, 2019.

\bibitem{li2021albef}
J. Li, R. R. Selvaraju, A. D. Gotmare, S. Joty, C. Xiong, and S. Hoi, 
``Align Before Fuse: Vision and Language Representation Learning with Momentum Distillation,'' 
\textit{arXiv preprint arXiv:2107.07651}, 2021. 
doi: 10.48550/arXiv.2107.07651.

\bibitem{li2022blip}
J. Li, D. Li, C. Xiong, and S. Hoi, 
``BLIP: Bootstrapping Language-Image Pre-training for Unified Vision-Language Understanding and Generation,'' 
\textit{arXiv preprint arXiv:2201.12086}, 2022. 
doi: 10.48550/arXiv.2201.12086.

\bibitem{kim2021vilt}
W. Kim, B. Son, and I. Kim, 
``ViLT: Vision-and-Language Transformer Without Convolution or Region Supervision,'' 
\textit{arXiv preprint arXiv:2102.03334}, 2021. 
doi: 10.48550/arXiv.2102.03334.

\bibitem{tan2019lxmert}
H. Tan and M. Bansal, 
``LXMERT: Learning Cross-Modality Encoder Representations from Transformers,'' 
\textit{arXiv preprint arXiv:1908.07490}, 2019. 
doi: 10.48550/arXiv.1908.07490.

\bibitem{chen2019uniter}
Y.-C. Chen, L. Li, L. Yu, A. El Kholy, F. Ahmed, Z. Gan, Y. Cheng, and J. Liu, 
``UNITER: UNiversal Image-TExt Representation Learning,'' 
\textit{arXiv preprint arXiv:1909.11740}, 2019. 
doi: 10.48550/arXiv.1909.11740.



\bibitem{vaswani2017attention}
A.~Vaswani \emph{et al.}, ``Attention is all you need,'' 
in \emph{Advances in Neural Information Processing Systems (NeurIPS)}, 2017.

\bibitem{qiskit}
H.~Abraham \emph{et al.}, ``Qiskit: An open-source framework for quantum computing,'' 
\emph{Zenodo}, 2019. [Online]. Available: https://doi.org/10.5281/zenodo.2562111

\bibitem{pennylane}
V.~Bergholm \emph{et al.}, ``PennyLane: Automatic differentiation of hybrid quantum--classical computations,'' 
\emph{arXiv preprint arXiv:1811.04968}, 2018.

\bibitem{biamonte2017quantum}
J.~Biamonte \emph{et al.}, ``Quantum machine learning,'' 
\emph{Nature}, vol. 549, no. 7671, pp. 195--202, 2017.

\bibitem{schuld2021ml}
M.~Schuld, ``Machine learning with quantum computers,'' 
\emph{Springer}, 2021.

\bibitem{raimond2001manipulating}
J. M. Raimond, M. Brune, and S. Haroche, 
``Manipulating quantum entanglement with atoms and photons in a cavity,'' 
\textit{Reviews of Modern Physics}, vol. 73, no. 3, pp. 565--582, Aug. 2001. 
doi: 10.1103/RevModPhys.73.565.

\bibitem{wootters1998entanglement}
W. K. Wootters, 
``Quantum entanglement as a quantifiable resource,'' 
\textit{Philosophical Transactions of the Royal Society A: Mathematical, Physical and Engineering Sciences}, 
vol. 356, no. 1743, pp. 1717--1731, 1998. 
doi: 10.1098/rsta.1998.0244.

\bibitem{tamilarasan2025qasvs}
C. Tamilarasan, J. P. Babu, P. K. Kumar, V. Keerthi, R. R. Hossein, and P. Putta, 
``Quantum-Assisted Adaptive Stereo Vision System (QASVS) for Autonomous Robotics,'' 
in \textit{Proc. 2025 Int. Conf. Metaverse and Current Trends in Computing (ICMCTC)}, 
Subang Jaya, Malaysia, 2025, pp. 1--4. 
doi: 10.1109/ICMCTC62214.2025.11196130.

\bibitem{havlivcek2019supervised}
V.~Havl{\'\i}{\v{c}}ek \emph{et al.}, ``Supervised learning with quantum-enhanced feature spaces,'' 
\emph{Nature}, vol. 567, no. 7747, pp. 209--212, 2019.

\bibitem{quantum_kernel}
M.~Schuld and N.~Killoran, ``Quantum machine learning in feature Hilbert spaces,'' 
\emph{Phys. Rev. Lett.}, vol. 122, no. 4, p. 040504, 2019.

\bibitem{boyapally2025qsannviq}
D. Boyapally, S. Alla, A. S. Sichani, S. M. A. Rahaman, and H. A. Akbarpour, 
``QSANN-ViQ: A Lightweight Quantum Self-Attention Vision Model for Explainable and Secure Edge Image Classification,'' 
in \textit{Proc. 2025 IEEE Int. Conf. Quantum Computing and Engineering (QCE)}, 
Albuquerque, NM, USA, 2025, pp. 622--623. 
doi: 10.1109/QCE65121.2025.10476.

\bibitem{qu2023qmfnd}
Z. Qu, Y. Meng, G. Muhammad, and P. Tiwari, 
``QMFND: A quantum multimodal fusion-based fake news detection model for social media,'' 
\textit{Information Fusion}, vol. 98, p. 102172, 2023. 
doi: 10.1016/j.inffus.2023.102172.

\bibitem{chen2024qmcl}
C.-S. Chen, A. H.-W. Tsai, and S.-C. Huang, 
``Quantum Multimodal Contrastive Learning Framework,'' 
\textit{arXiv preprint arXiv:2408.13919}, 2024. 
doi: 10.48550/arXiv.2408.13919.

\bibitem{lin2014coco}
T.-Y. Lin \textit{et al.}, 
``Microsoft COCO: Common Objects in Context,'' 
\textit{arXiv preprint arXiv:1405.0312}, 2014. 
doi: 10.48550/arXiv.1405.0312.

\bibitem{young2014image}
P. Young, A. Lai, M. Hodosh, and J. Hockenmaier, 
``From Image Descriptions to Visual Denotations: New Similarity Metrics for Semantic Inference over Event Descriptions,'' 
\textit{Trans. Assoc. Comput. Linguistics}, vol. 2, pp. 67--78, 2014.

\bibitem{mcclean2018barren}
J. R. McClean, S. Boixo, V. N. Smelyanskiy, R. Babbush, and H. Neven, 
``Barren Plateaus in Quantum Neural Network Training Landscapes,'' 
\textit{arXiv preprint arXiv:1803.11173}, 2018. 
doi: 10.48550/arXiv.1803.11173.

\bibitem{ibmq_noise}
A.~Kandala \emph{et al.}, ``Error mitigation extends the computational reach of a noisy quantum processor,'' 
\emph{Nature}, vol. 567, pp. 491--495, 2019.

\bibitem{oord2018cpc}
A. van den Oord, Y. Li, and O. Vinyals, 
``Representation Learning with Contrastive Predictive Coding,'' 
\textit{arXiv preprint arXiv:1807.03748}, 2018. 
doi: 10.48550/arXiv.1807.03748.

\bibitem{hasan2023qcs}
S. R. Hasan, M. Z. Chowdhury, M. Sayem, and Y. M. Jang, 
``Quantum Communication Systems: Vision, Protocols, Applications, and Challenges,'' 
\textit{IEEE Access}, vol. 11, pp. 15855--15877, 2023. 
doi: 10.1109/ACCESS.2023.3244395.


\bibitem{huang2020predicting}
H.-Y.~Huang, R.~Kueng, and J.~Preskill,
``Predicting many properties of a quantum system from very few measurements,''
\emph{Nature Physics}, vol.~16, pp.~1050--1057, 2020,
doi: 10.1038/s41567-020-0932-7.


\bibitem{cerezo2021cost}
M.~Cerezo, A.~Sone, T.~Volkoff, L.~Cincio, and P.~J.~Coles,
``Cost function dependent barren plateaus in shallow parametrized
quantum circuits,''
\emph{Nature Communications}, vol.~12, Art.~no.~1791, 2021,
doi: 10.1038/s41467-021-21728-w.


\bibitem{quantum_advantage_ml}
E.~Tang, ``A quantum-inspired classical algorithm for recommendation systems,'' 
in \emph{Proc. ACM Symp. Theory of Computing (STOC)}, 2019.

\end{thebibliography}
\end{document}